# ZeShot-VQA: Zero-Shot Visual Question Answering Framework with Answer Mapping for Natural Disaster Damage Assessment

Ehsan Karimi
*Lehigh Uniersity*
Pennsylvania, USA
ehk224@lehigh.edu

Maryam Rahnemoonfar[*]
*Lehigh Uniersity*
Pennsylvania, USA
maryam@lehigh.edu
[*]Corresponding author

*Abstract*—Natural disasters usually affect vast areas and devastate infrastructures. Performing a timely and efficient response is crucial to minimize the impact on affected communities, and data-driven approaches are the best choice. Visual question answering (VQA) models help management teams to achieve in-depth understanding of damages. However, recently published models do not possess the ability to answer open-ended questions and only select the best answer among a predefined list of answers. If we want to ask questions with new additional possible answers that do not exist in the predefined list, the model needs to be fin-tuned/retrained on a new collected and annotated dataset, which is a time-consuming procedure. In recent years, large-scale Vision-Language Models (VLMs) have earned significant attention. These models are trained on extensive datasets and demonstrate strong performance on both unimodal and multi-modal vision/language downstream tasks, often without the need for fine-tuning. In this paper, we propose a VLM-based zero-shot VQA (ZeShot-VQA) method, and investigate the performance of on post-disaster FloodNet dataset. Since the proposed method takes advantage of zero-shot learning, it can be applied on new datasets without fine-tuning. In addition, ZeShot-VQA is able to process and generate answers that has been not seen during the training procedure, which demonstrates its flexibility.

## I. Introduction

FOLLOWING a natural disaster, rapid response teams must assess the extent of the damage, identify affected areas in need of immediate assistance, and prioritize their actions. The success of the decisions made during this phase is heavily reliant on the speed and precision of the data collected from the field. Traditional methods of data gathering, such as field surveys and social media monitoring, are often time-consuming [1] and insufficient. Aerial imagery, which provides extensive coverage and detailed information, is considered one of the most valuable data sources. As a result, technologies like satellites and unmanned aerial vehicles (UAVs) have gained significant attention in recent years [2]–[4]. However, due to the high cost and technical limitations associated with satellite imagery—such as issues with object occlusion, image resolution, and restricted viewing angles—the use of UAVs has grown substantially [5]. The expanded availability of aerial data has facilitated the development of image processing algorithms designed to automate and expedite the generation of reports for estimating the extent of damage using such images.

Visual Question Answering (VQA) is an interdisciplinary task that combines image processing and natural language processing. Unlike image captioning, which provides a general description of an image's content, VQA models focus on extracting specific text-aware local features and details of an image in order to answer questions. This allows for the retrieval of relevant high-level information by posing precise questions [6]. This capability of VQAs assists disaster response teams in analyzing UAV-captured images automatically, enabling them to extract valuable information for post-disaster management and decision-making.

Existing studies focus on training and fine-tuning VQA models with datasets from natural disasters and remote sensing. The frameworks proposed by these studies suffer from two main drawbacks: 1) Models can only answer questions if the potential answers are included in a predefined list of answers established during the training or fine-tuning stage [7]–[10]. 2) Natural disaster datasets are usually collected following specific catastrophic events, often focusing on a geographically limited area impacted by the disaster. Due to variations in natural environments, rural and urban structures, and vegetation across different locations, models trained on a dataset may face challenges in generalizing when applied to other regions. These limitations highlight the need for a VQA framework that can generate previously unseen answers and demonstrate strong generalization capabilities [9].

Recently developed Large-scale Vision-Language Pretraining (VLP) frameworks, such as CLIP [11], ALIGN [12], and BLIP [13] have shown remarkable generalization abilities. In addition, VQA models built on these frameworks are capable of answering open-ended questions which alleviate the first challenge [14]. However, even these advanced models struggle to generate accurate answers when applied to natural disaster datasets. This limitation arises from the gap between the general-purpose datasets used during the pretraining phase and the specific characteristics of disaster-related data.

In this study, we propose a zero-shot framework inspired by large-scale vision-language models to address the challenges outlined above. The proposed framework is capable of answering novel questions, even when the possible answers are not included in the predefined answer list. In addition, proposed framework benefits from the generalization capabilities of large-scale vision-language models. Moreover, this approach can be applied to data from a new natural disaster without the need for a newly annotated dataset from the affected location for training or fine-tuning. This characteristic not only accelerates the decision-making process but also enhances the efficiency of rescue operations.

The remainder of this paper is structured as follows. Section II presents a brief review of related works, discussing their advantages and limitations. Section III introduce dateset employed in this study, and section IV explains the methodology and framework pipeline along with presenting the experiments and results on FloodNet dataset. Final section offers the conclusion.

## II. RELATED WORKS

Almost all recent VQA methods in remote sensing and natural disasters use classification objectives to find the answer of the input question regarding the given image [8], [15]. To achieve this, all possible answers are considered as classes, and the model predicts the closest class as the answer. RSVQA [16] creates two remote sensing classification datasets based on OpenStreetMap. CNN and RNN deep-learning methods are adopted in this research to extract features of given image and question. Sequentially, point-wise multiplication of extracted feature representations is used for answer prediction. [17] separates categorical and counting tasks outputs as a classification and regression tasks respectively. RSVQA also categorizes the counting answers to alleviate the data imbalance and facilitate the task. LIT-4-RSVQA [18] is built upon transformer-based models, $BERT_{TINY}$ and $XCiT_{Nano}$, to generate embedding vectors for the given image and the question. After concatenating these vectors and passing them through the MLP-based fusion layers, a soft-max function is applied to select the best answer among possible answers. The model proposed by [19] Follows the same approach while using the Bag of Words (BoW) to represent the question in post-disaster datasets. Introduced model in [20] is the modified version of the RSVQA that investigates the effect of replacing the RNN with BERT or Skip-thought models to extract features from the input text. VQA framework proposed in [21] uses a multi-class classification model to assign labels to the input image, and in the next step, fed a language model by the question along with these labels for feature extraction and answer selection using a classifier. studies [22] and [15] adopt cross-attention mechanism to extract text-aware visual feature extraction and image-aware textual feature extraction before information fusion and classification. The method presented by [23] focuses on improving the performance of the model in counting tasks on FloodNet [2] dataset by exploiting CapsulNet [24] and transformers. The approach proposed by [25] introduces cross-modal global attention to enhance text-aware visual feature extraction and incorporates the SPCL framework [26] to initiate model training with easier questions, progressively incorporating more difficult ones as training progresses.

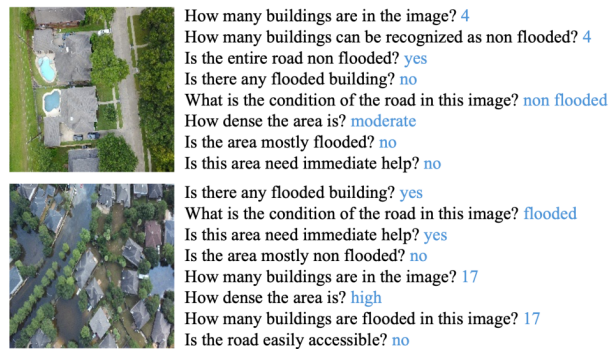

Fig. 1: Image-question-answer triplet samples in FloodNet.

One factor contributing to the higher accuracy in classification-based approaches is the limited set of possible answers, which inadvertently improves the accuracy of these VQA models. Specifically, this approach can result in information leakage between the training and testing stages, particularly in counting tasks, because the output classes are defined based on all possible answers and only include acceptable predefined numbers. Furthermore, counting questions are particularly challenging, especially within the FloodNet dataset [10], [17]. The classification approach often struggles to provide accurate answers due to two main issues: 1) the significant imbalance in the distribution of counting classes, particularly in FloodNet, and 2) the presence of unseen numbers during training, which reduces model performance in real-world scenarios [2], [7], [10], [27], [28]. To address these challenges, RSVQA transforms the regression task into a classification problem by grouping numbers into quartiles. In contrast, the methods proposed by [28] and [17] treat this problem as an open-ended and regression task, respectively. [17] employ root mean squared error (RMSE) to predict exact numerical values. In Zero-Shot VQA techniques approaches the task as a classification problem with an infinite number of possible classes. Authors of [9] generate a caption for the given image and then use a large language model (LLM) to answer the question based on that caption. However, this approach loses significant information by transmitting image details solely through the caption. In contrast, this study, ZeShot-VQA, directly incorporates image information into the answer generation process, ensuring that one of the possible choices is selected and thereby reducing the occurrence of False Negative answers.

## III. DATA

Experiments are conducted using the FloodNet dataset, which was collected following Hurricane Harvey in 2017 by an unmanned aerial vehicle (UAV) over Fort Bend County, Texas. The FloodNet dataset consists of 2,188 images representing both flooded and non-flooded areas, along with 7,355 paired questions-answers. These questions are based on 31 unique queries, which are organized into seven sub-categories and four main categories. Fig.1 demonstrates samples from the FloodNet.

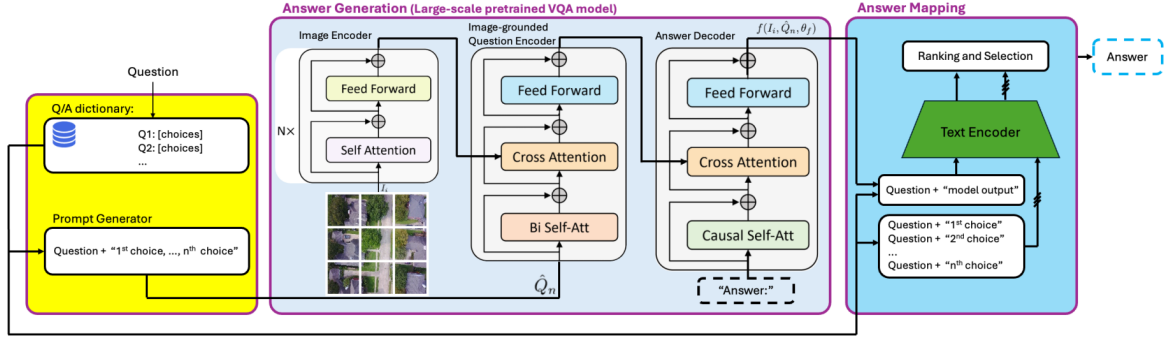

Fig. 2: The proposed ZeShot-VQA framework for zero-shot question answering.

*a) Counting:* questions can be categorized into two main types: *complex counting* and *simple counting*. Complex counting questions focus on determining the number of buildings based on specific conditions, such as whether they are flooded or not, as in the question "How many flooded buildings are visible in this image?". In contrast, simple counting questions inquire about the total number of buildings, irrespective of their condition, like "What is the total number of buildings?". Both types of questions are typically framed as regression tasks.

*b) Condition recognition:* questions are designed to assess the status of areas and objects in an image, specifically determining whether they are flooded or not. These questions are typically categorized into two types: yes/no questions and multiple-choice questions. Examples include "What is the condition of the road?", "What is the overall condition of the given image?", and "Is there any flooded building?". Possible answers to these questions can include categories such as partially flooded, non-flooded, flooded, non-flooded, flooded, and yes, no, respectively.

*c) Density estimation:* questions aim to assess the level of density within an area, with possible responses as low, high, or moderate.

*d) Risk assessment:* questions are typically framed as yes/no inquiries, intended to evaluate whether an area requires urgent intervention.

## IV. METHODOLOGY

The scarcity of large, human-annotated datasets in the field of natural disasters makes it challenging to train high-performance models with strong generalization for this application. Leveraging pretrained VLMs with zero-shot learning presents a promising solution to address this issue. However, because of the gap between the general datasets used for VLMs pretraining and natural disaster datasets, these models fail in providing accurate answers. In this section, we introduce the proposed transformer-based method, ZeShot-VQA framework, which uses the large-scale pretrained VQA models derived from pretrained VLMs, for question answering, and fill the mentioned gap by prompt engineering and adopting language models.

### A. Framework Architecture

Fig. 2 illustrates our framework, which consists of three main stages as follows.

*1) Prompt Modification:* Since pretrained large-scale VQA models tend to generate more general answers, we modify the questions and embed additional context to guide the model toward producing more relevant responses. In this stage, "yes/no" and multiple-choice questions are modified based on a designed format, before being input into the model. We created a dictionary of all unique questions and their possible answers for multiple-choice and yes/no questions $QA = \{Q_1 : A_1 = \{A_1^1, A_1^2, ..., A_1^{M_1}\}, Q_2 : A_2 = \{A_2^1, A_2^2, ..., A_2^{M_2}\}, ..., Q_N : A_N = \{A_N^1, A_N^2, ..., A_N^{M_N}\}\}$. Where $Q_N$ is the $N^{th}$ unique question, and $A_N^{M_N}$ is the $M^{th}$ possible answer of $Q_N$. For each input question, we concatenated all possible answer choices with a comma, and then the combined string is appended to the question. For example, the question *"What is the current state of the area?"* is transformed to *"What is the current state of the area? non-flooded, flooded"*. Simple and complex counting questions, however, remain unchanged.

*2) Answer Generation:* In this stage, a pretrained general-purpose VQA model $f$ is used to generate answers for modified questions $\hat{Q}_n$ based on the provided image $I_i$ and weights $\theta_f$. Specifically, in this work, we develop our general-purpose VQA model based on the BLIP VLM, as shown in Fig. 2. ZeShot-VQA encodes the input image $I_i$ using the BLIP image encoder and shares the resulting feature vector with the text encoder to extract image-grounded features from the given modified question $\hat{Q}_n$. The output of the Image-grounded Question Encoder block is then used to generate the raw answer $f(I_i, \hat{Q}_n, \theta_f)$ using Answer Decoder block.

*3) Answer Matching:* In this stage, we formulate a matching problem where the answer generated by the general-purpose VQA model ($f(I_i, \hat{Q}_n, \theta_f)$) must be mapped to one of the available choices in multiple-choice or yes/no questions ($A_n^m$). This matching process helps the framework to fill a part of the gap between general datasets used for pretrainig and the natural disaster datasets. To identify the most relevant choice conceptually, we apply a 3-step mapping process. In the first step we create a query set from the given question $Q_n$ separately concatenated with all possible answers $QA_n = \{QA_n^1 = Q_n + A_n^1, QA_n^2 = Q_n + A_n^2, ..., QA_n^M = Q_n + A_n^M\}$. A reference query is also created by concatenating $Q_n$ and raw generated answer $f(I_i, \hat{Q}_n, \theta_f)$ named $Qf_{i,n}$. Sequentially, language model (CLIP text encoder) $g$ is applied to calculate the feature representations of queries along with reference query. Finally, the cosine similarity scores between the reference query and query set items are calculated using

$$Sim(A_n^m, f(I_i, \hat{Q}_n, \theta_f)) = \frac{g(Qf_{i,n}, \theta_g) \cdot g(QA_n^m, \theta_g)}{|g(Qf_{i,n}, \theta_g)| \cdot |g(QA_n^m, \theta_g)|} \quad (1)$$

and $A_n^m$ with maximum similarity score is selected as the output answer.

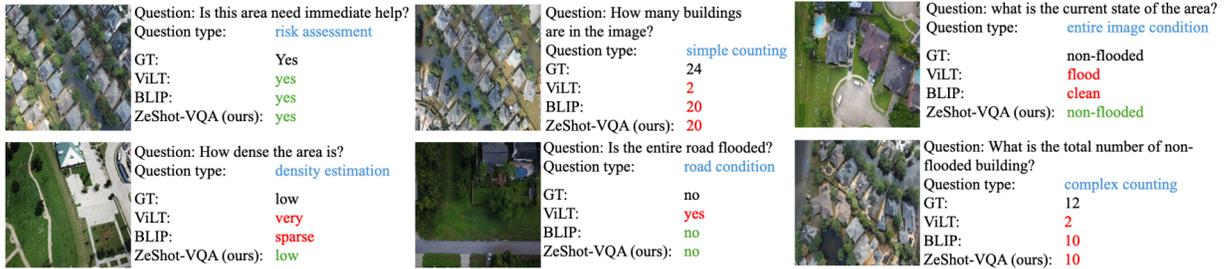

Fig. 3: Example predictions made by ViLT, BLIP, and ZeShot-VQA (ours) for sample questions from FloodNet

TABLE I: VQA ACCURACY (%) COMPARISON ON THE FLOODNET DATASET ACROSS DIFFERENT MODELS.

| Question Type | ViLT [29] | BLIP [13] | BLIP-2 [30] | Qwen2-VL [31] | ZeShot-VQA (ours) |
|---|---|---|---|---|---|
| Building Condition | 51.91 | **78.14** | 40.98 | 53.00 | **78.14** |
| Complex Counting | 17.00 | **22.50** | 12.56 | 16.50 | **22.50** |
| Density Estimation | 22.40 | 23.49 | 26.78 | 36.12 | **38.79** |
| Entire Condition | 30.90 | 46.47 | 58.39 | 56.69 | **59.12** |
| Risk Assessment | 24.04 | **88.52** | 28.41 | 85.24 | **88.52** |
| Road Condition | 27.14 | 34.08 | 28.77 | **71.4** | 40.53 |
| Simple Counting | 18.57 | **27.32** | 13.66 | 9.29 | **27.32** |

*B. Results and Discussion*

Table I presents a comparison of the performance of the proposed ZeShot-VQA framework with the zero-shot evaluations of BLIP and ViLT. All values are expressed as percentages, with accuracy as the evaluation metric. As discussed in Section II, CLIP text encoder is used to generate queries' representation vectors in the answer matcher block. Finally, cosine similarity is considered as the metric to find the closest choice to the output of the answer generator.

The results indicate that BLIP outperforms ViLT across all tasks, irrespective of the question type, and the proposed ZeShot-VQA outperforms BLIP, BLIP-2, and Qwen2-VL. Furthermore, the overall performance of the framework is heavily dependent on the performance and generalization abilities of the base models. Given that the zero-shot technique is utilized throughout all stages, it is possible to substitute any component with a higher-performing model to potentially improve the framework's effectiveness (see Fig. 3).

It is evident that all models struggle with counting tasks, which yield the lowest performance across all tasks. The primary reason for this is that counting is one of the most challenging tasks in VQA, particularly in remote sensing datasets [28]. In the density estimation task, the proposed model with answer mapping demonstrates significantly superior performance compared to ViLT and BLIP, highlighting the gap between datasets from a literature perspective. For example, in several instances, the model's outputs were classified as 'scarce', while the acceptable answers were 'low', 'moderate', and 'high'. The answer mapping model is employed to bridge this gap by selecting 'low' as the correct answer. It is important to note that the definitions of low, moderate, and high density within the dataset also contribute to this low accuracy. As a result, directly modifying the question or rephrasing it from a different perspective could help to enhance the model's performance. For instance, when the question is "How dense is the area?", it may not be clear whether the query pertains to vegetation density or building density, particularly when using zero-shot learning with a model trained on a general-purpose dataset without background knowledge in natural disaster. Furthermore, even if the question is clarified, it remains necessary to define the thresholds that separate the categories of density. We aim to address these issues in future studies.

## V. CONCLUSION

In this paper, we leveraged large-scale pretrained vision-language models (VLMs) and proposed a zero-shot visual question answering framework for post-disaster management applications. Specifically, BLIP is employed to generate general answers using a zero-shot approach, and answers are subsequently refined by CLIP to match the model's output with the closest choice among the acceptable choices. In contrast to other research, the proposed framework is not restricted to predefined answers during the information extraction phase, allowing any image-question-answer triplet to be used in the inference process to extract new information from visual data.